\documentclass[conference]{IEEEtran}
\IEEEoverridecommandlockouts
\usepackage{cite}
\usepackage{amsmath,amssymb,amsfonts}
\usepackage{algorithmic}
\usepackage{graphicx}
\usepackage{textcomp}
\usepackage{threeparttable}
\usepackage{amsmath,amsfonts}
\usepackage{algorithmic}
\usepackage{algorithm}
\usepackage{array}
\usepackage{textcomp}
\usepackage{stfloats}
\usepackage{url}
\usepackage{bbding} 
\usepackage{verbatim}
\usepackage{graphicx}
\usepackage{cite}
\usepackage{multirow}
\usepackage{pifont}
\usepackage{amssymb}
\usepackage{tikz}
\usepackage{fontawesome}
\usepackage{subfigure}
\usepackage{xcolor}
\usepackage{url}
\usepackage{booktabs}
\def\BibTeX{{\rm B\kern-.05em{\sc i\kern-.025em b}\kern-.08em
    T\kern-.1667em\lower.7ex\hbox{E}\kern-.125emX}}
\begin{document}
\title{Masked EEG Modeling for Driving Intention Prediction\\

}

\author{\IEEEauthorblockN{Jinzhao Zhou, Justin Sia, Yiqun Duan, Yu-Cheng Chang, Yu-Kai Wang, Chin-Teng Lin}
\IEEEauthorblockA{\textit{Faculty of Engineering and Information Technology} \\
\textit{University of Technology Sydney}\\
Ultimo, Australia \\
\{jinzhao.zhou, justin.sia, yiqun.duan, yu-cheng.chang, yukai.wang@uts.edu.au, chin-teng.lin@uts.edu.au\}@student.uts.edu.au}
}

\maketitle

\begin{abstract}
Driving under drowsy conditions significantly escalates the risk of vehicular accidents. 
Although recent efforts have focused on using electroencephalography (EEG) to detect drowsiness, helping prevent accidents caused by driving in such states, seamless human-machine interaction in driving scenarios requires a more versatile EEG-based system. 
This system should be capable of understanding a driver's intention while demonstrating resilience to artifacts induced by sudden movements.
This paper pioneers a novel research direction in BCI-assisted driving, studying the neural patterns related to driving intentions and presenting a novel method for driving intention prediction (DIP). 
In particular, our preliminary analysis of the EEG signal using independent component analysis (ICA) suggests a close relation between the intention of driving maneuvers and the neural activities in central-frontal and parietal areas. Power spectral density analysis at a group level also reveals a notable distinction among various driving intentions in the frequency domain. To exploit these brain dynamics, we propose a novel Masked EEG Modeling (MEM) framework for predicting human driving intentions, including the intention for left turning, right turning, and straight proceeding. 
Extensive experiments, encompassing comprehensive quantitative and qualitative assessments on a publicly available driving dataset, demonstrate the proposed MEM method is proficient in predicting driving intentions across various vigilance states. Specifically, our model attains an accuracy of $85.19\%$ when predicting driving intentions for drowsy subjects, which shows its promising potential for mitigating traffic accidents related to drowsy driving. Ablation also demonstrates that our method significantly enhances the flexibility in handling missing channels. Notably, our method maintains over $75\%$ accuracy when more than half of the channels are missing or corrupted, underscoring its adaptability in real-life driving scenarios.
\end{abstract}


\section{Introduction}
Drowsy driving poses a substantial and increasing risk to transportation safety. 
In a drowsy cognitive state, a driver experiences impairment in both physical and mental faculties, leading to substantial delays in critical responses necessary to prevent accidents~\cite{mohammedi2023methods, rahman2023understanding}. To ensure safe driving, Brain-Computer Interface (BCI) technology has emerged as a precaution through real-time monitoring of the driver's cognitive state \cite{reddy2021eeg}. However, in the middle of long-distance driving, understanding and interpreting a driver's intentions in moments preceding potential danger can provide dynamic and nuanced safeguards that are aligned with the driver's objectives. Furthermore, understanding the driver's intentions could mark a pivotal stride in the evolution of BCI-assisted driving and the integration of human intention into the realm of human-machine interactions research.

In recent years, significant progress has been made in the decoding of tele-command \cite{palumbo2021motor}, motor imagery (MI) \cite{zhou2023speech2eeg,duan2023cross}, and language \cite{duan2023dewave} from non-invasive electroencephalography (EEG) signals. These studies have the potential to help people restore motor functionalities, control external devices, and reestablish communication with the outside world \cite{alabboudi2020eeg,nieto2022thinking,bouton2016restoring}. Nonetheless, extracting a driver's intention from neural signals before the action is a considerable challenge. On the one hand, action planning and decision-making often involve complex cognitive processes. 
On the other hand, dissimilar to tasks such as MI, which typically allows two or more seconds for distinctive brain patterns to emerge from a tranquility state \cite{altaheri2023deep}, predicting the driver's intention necessitates working with a brief EEG signal duration, potentially limiting the clarity of discerning underlying neural patterns associated with driving intentions \cite{murphey2022past,xu2021review,chang2022decoding}. 
Last but not least, the proposed driving scenario is different from the controlled settings common in most EEG studies due to the dynamic driving environment, inherently posing challenges of sudden movements and potential electrode connectivity loss.
Regrettably, current EEG decoding methods often falter when faced with heavily noisy signals or compromised electrodes, as highlighted in existing literature \cite{kuang2021cross,banville2022robust}.

In this paper, we study the brain activities associated with driving intentions by comprehensive analysis of decomposed EEG components, elucidating the relevant EEG channels and frequency bands associated with driving decisions. This analysis not only enhances the understanding of the neuro mechanisms involved but also provides insights for channel and frequency selection to remove inference from unrelated brain activities. For the challenge of modeling and predicting intention from EEG signals, we introduce a mask modeling approach called Masked EEG Modeling (MEM), aiming to improve the semantic level of EEG representation through a masked reconstructive training objective. To bolster the model's resilience against corrupted information or missing channels, two specialized masking strategies are incorporated in the frequency dimension and the channel dimension respectively. For the prediction of the driver's intention, we build a classifier utilizing the outputs from the MEM encoder. The contributions of this paper can be summarized as follows:
\begin{itemize}
    \item A meticulous analysis of EEG signals is provided that offers insights into the spatial dynamics of brain activity for driving intentions.
    \item A pioneering task-specific EEG framework is introduced to predict the driver's intentions, discerning between three directional commands including proceeding straight, turning right, and turning left.
    \item Two distinct masking techniques aimed at improving the learned EEG representations and the resilience against loss of information or channels are propose for EEG decoding research.
    \item Extensive experiments were carried out to furnish both quantitative and qualitative evaluations of the driving intention prediction models proposed herein.
\end{itemize}

\section{Related Works}
\subsection{EEG-based applications in driving scenarios}
Current BCI-based approaches related to our works are the EEG-based wheelchair control system~\cite{pinheiro2018eeg,ofner2017upper}. However, most approaches rely on the steady-state visual evoked potential (SSVEP) or MI methods which are not ideal to be applied in driving scenarios \cite{al2018review}. For instance, the visual stimuli for the SSVEP approach could pose significant dangers as additional distractions to the driver while the MI approach limits users to envisioning limb movements that are unnatural during driving. 

On the other hand, existing research associating the human brain with driving mainly focused on estimating the driver's vigilance state \cite{lin2005estimating,guo2017detection}, fatigue \cite{hu2018automated} or mental workload \cite{di2018eeg}. To the best of our knowledge, no prior studies have investigated brain activities for the prediction of driving maneuver intentions.
\subsection{Deep learning methods for EEG decoding}
To decode driving intentions efficiently, various deep learning models have been proposed for effective representations learning from EEG signals. 
Widely recognized networks such as EEGNet \cite{lawhern2018eegnet} and TCNet \cite{ingolfsson2020eeg} leverage temporal and spatial features through convolutional neural networks (CNNs). 
Diverging from pure CNN architectures, Long Short-Term Memory (LSTM) models have been employed to enhance recognition performance on extended EEG signals \cite{xu2020one}. 
Generalized architectures, such as Transformers \cite{luo2023shallow,ma2022novel}, have been implemented to enhance the model's capability for representation learning. Despite their proficiency in capturing temporal and spatial information, these models lack flexibility in handling missing or corrupted channels. Leveraging the efficacy of the recent masked modeling approach in diverse tasks \cite{huang2022masked}, our work extends this strategy into the EEG domain. We introduce tailored masking strategies that enable effective EEG representation learning while accommodating missing information or channels.

\section{Method}\label{sec:method}
In this section, we delve into the details of the proposed MEM method for driving intention prediction. We begin with a concise overview of the DIP task in Section \ref{Sec:def}. and subsequently, introduce the MEM method in Section \ref{sec:mae}.

\subsection{Task Setting for Driving Intention Prediction} \label{Sec:def}
\begin{figure}[!ht]
    \centering
    \includegraphics[width=0.95\columnwidth]{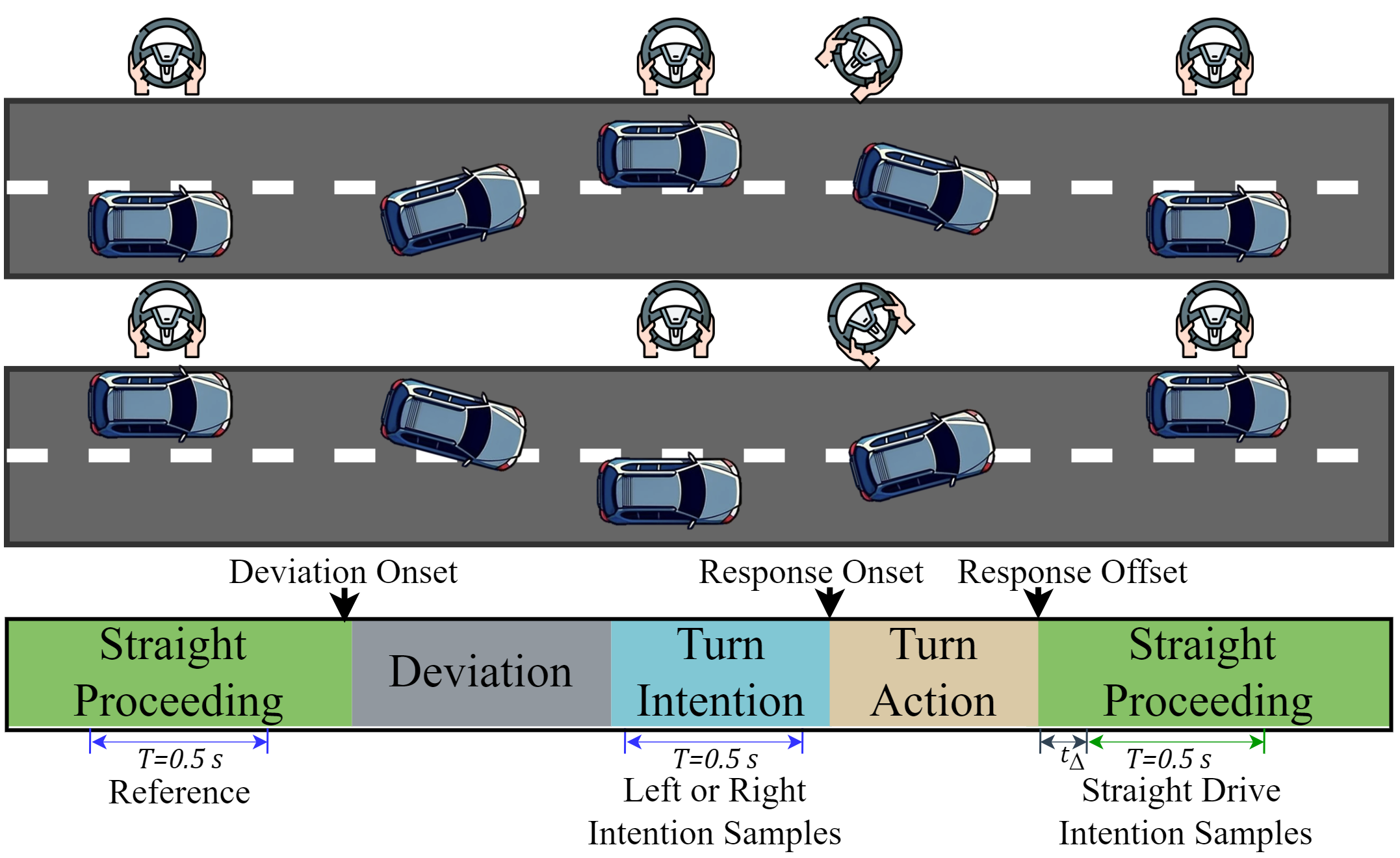}
    \caption{Illustration of the driving intention prediction task. The driver starts from straight proceeding on the road before the deviation occurs at a random time (deviation onset). After the deviation onset, the driver needs to take action to steer back to the original lane. Finally, the car is back in the original lane and the driver focuses on driving straight.}
    \label{fig:driving-scenario}
\end{figure}

The driving scenario for the DIP task is depicted in Figure \ref{fig:driving-scenario}. During driving, the car may deviate from the lane, requiring the driver to take corrective action. For real-time applicability and the reliability of neural patterns for driving intention, we extract EEG signals of duration $T=0.5$ seconds before the action onset as sample data for left or right turning intentions. Similarly, EEG signals for straight proceeding intention are obtained $T+t_\Delta$ seconds after the completion of the driving action, where $t_\Delta$ is a $0.1$-second offset. Conversely, the period of straight proceeding before deviation onset serves as the reference signal for mitigating inter-subject variations.

Given an EEG interface with a sampling rate of $f_s$, we are provided with a segment of multi-channel EEG signal $\mathbf{e}$ consisting of $L=\!T\!\times\!f_s$ timesteps and $N$ channels. The objective of the DIP task is to forecast the driving intention, denoted by $\hat{c}$ solely based on $\mathbf{e}$. Hence, the objective is to maximize the probability of the predicted driver intention $p(\hat{c}|\mathbf{e})$.

\begin{figure*}
    \centering
    \includegraphics[width=1.0\textwidth]{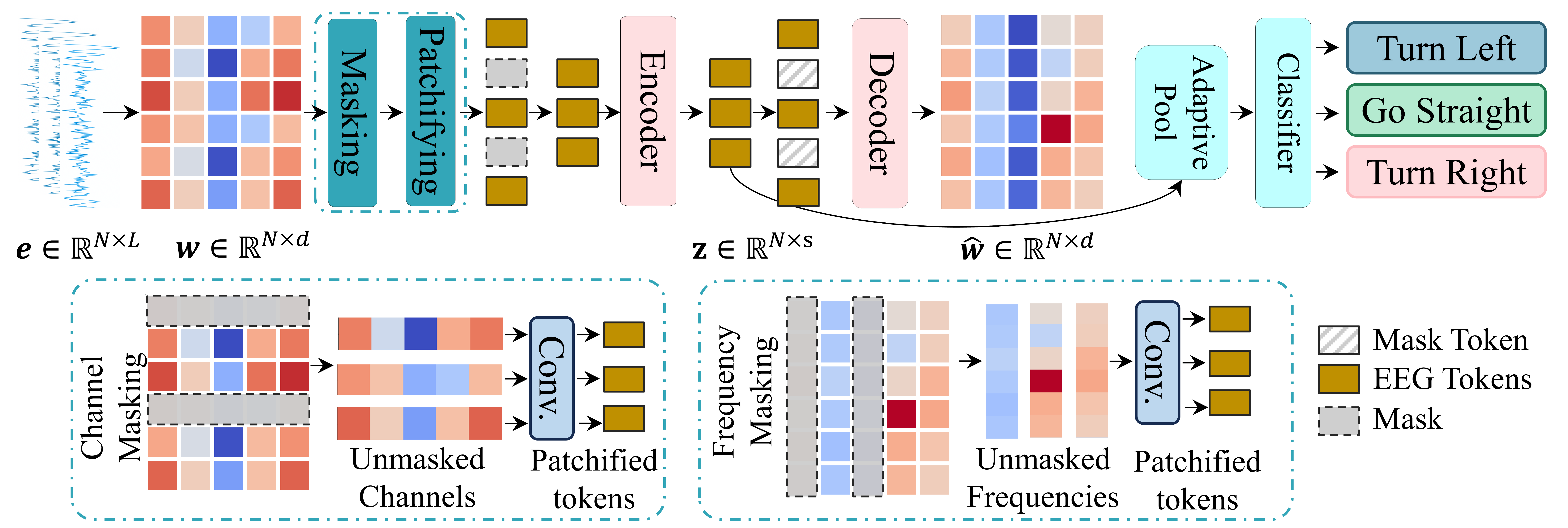}
    \caption{Illustration of the model structure, the EEG waves $\mathbf{e}$ are first converted into frequency spectrogram $\mathbf{w}$ using Welch's method. Afterward, we patchify $\mathbf{w}$ and perform random masking to the EEG tokens using channel masking or frequency masking strategies. Subsequently, a transformer-based encoder processes the unmasked EEG tokens into the latent representation $\mathbf{z}$. Then, a decoder is used to reconstruct the spectrogram $\hat{\mathbf{w}}$ using the latent representations and the mask tokens. For intention prediction, a classifier is used to classify the latent representations into different steering directions. \label{fig:modelstructure} }
\end{figure*}

\subsection{Masked EEG Modeling} \label{sec:mae}
We introduce the MEM method, which leverages the masked autoencoder (MAE) \cite{he2022masked} as the foundational model for effective representation learning in DIP. The integration of MAE is mainly used to address the challenges inherent in learning high-quality EEG representations. Remarkably, the encoder-decoder architecture and the self-attention building blocks enable flexible exploitation of spatial or temporal dependencies from the EEG signal, which is crucial for acquiring robust EEG filters and latent representations. Notably, the reconstructive self-supervised learning scheme compels the encoder to capture essential information from the input signal, enhancing the overall semantic level of the learned EEG representation.

The proposed MEM architecture is illustrated in Figure~\ref{fig:modelstructure}. After frequency domain conversion, our MEM method transforms the EEG spectrogram into EEG tokens by the patchifying operation. Subsequently, a masking strategy is applied to shuffle and mask out a portion of the unmasked EEG tokens. A lightweight transformer encoder is then utilized to process the unmasked tokens into latent EEG representations. Afterward, we use mask tokens that contain no information to represent missing tokens and fill the latent representation to the full length of the unmasked EEG tokens. Finally, another transformer decoder that shares the same structure as the encoder is used to reconstruct the original EEG signal from the latent EEG representation and the mask tokens. 

Compared to images, EEG signals differ significantly in terms of how their information is organized. While information from images is typically conveyed by patches that capture spatial patterns for visual clues, neural activities from EEG signals are mainly organized by channels and frequency bands. Therefore, patchifying and masking methods specific to EEG modeling are needed to learn effective representation learning. Moreover, to cope with the driving scenario, where drastic physical movements or faulty electrodes often introduce noise or cause channel disconnection, we present two specifically designed masking strategies in Section \ref{subsec:channel-masking} and \ref{subsec:frequency-masking}.

\paragraph{Preprocessing}
Utilizing the raw EEG signal $\mathbf{e}\in\mathbb{R}^{N\times{L}}$, we employ Welch's method~\cite{welch1967use} to derive the estimated Power Spectral Density (PSD), transforming the raw EEG waves into the frequency domain. This process involves dividing the EEG signal into overlapping subsections, each subjected to windowing using a Hann window to mitigate edge effects. Following this, a Fast Fourier Transform (FFT) is applied to each subsection, producing a set of frequency components for each segment. The power spectrum for each segment is then computed by squaring the magnitude of the Fourier coefficients. Finally, Welch's PSD estimation $\mathbf{w}\in\mathbb{R}^{N\times{d}}$ is a frequency spectrogram derived from the averaged power spectra of various segments, where $d$ represents the number of selected frequency components used to represent $\mathbf{e}$ in the frequency domain. This preprocessing step allowed us to estimate the PSD of each frequency component, capturing the energy distribution of the EEG signals at specific frequencies.

\paragraph{Channel masking}\label{subsec:channel-masking}
For the channel masking strategy, we divide the frequency spectrogram $\mathbf{w}$ into non-overlapping channels $\{\mathbf{w}_{i\cdot}\}_{i=1,\cdots, N}, \mathbf{w}_{i\cdot} \in \mathbb{R}^{d}$. Then we use a convolutional layer to process the frequency components from each channel into 1D EEG tokens of size $s$. Here, the convolutional layer works as a learnable filter in the frequency domain. Then we uniformly sample a number of channels to be masked out. This creates a self-supervised learning task for the encoder to infer complete global information from the partial observation of the EEG signal as well as learning the channel-wise spatial dependencies inherent in the EEG signal. More importantly, the EEG encoder can learn to handle different numbers of EEG channels, granting the model with flexibility to handle missing channels in real-life driving applications. 

\paragraph{Frequency Masking}\label{subsec:frequency-masking}
Diverging from the channel masking strategy, the frequency masking approach involves extracting and masking tokens along the frequency dimension by splitting \(\mathbf{w}\) into non-overlapping frequency bands across all channels \(\{\mathbf{w}_{\cdot{j}}\}_{j=1,\cdots, d}, \mathbf{w}_{\cdot{j}} \in \mathbb{R}^{N}\). We also employ a convolutional layer to transform each EEG patch into EEG tokens of size \(s\). Unlike the channel masking strategy, the convolutional layer mainly functions as a spatial filter adept at learning patterns specific to a given frequency band across all channels. The self-supervised learning task introduced by the frequency masking strategy differs markedly from that associated with the channel masking strategy, as it cannot be easily resolved through interpolation from unmasked channels. Hence, the encoder needs to make use of redundant or complementary information from the remaining frequency bands to mitigate the impact of the missing information or learn to dynamically adjust its sensitivity to other frequency bands when the most informative ones are masked out.

\paragraph{EEG Encoder and Decoder}
The MEM architecture utilizes $2$ transformer blocks with the same hidden embedding size $s$ as the EEG token for both the encoder and decoder. We use $s=512$ in our model. Each transformer block is configured with $4$ attention heads. To ensure consistency in input length, mask tokens are incorporated into the latent representations $\mathbf{z}\in\mathbb{R}^{N\times{s}}$ by the encoder, aligning the input to the decoder with the original EEG tokens' length before masking. Additionally, positional embeddings are added to the input EEG tokens before masking and the full-size latent representation before being fed to the decoder.

\paragraph{Classifier}
To utilize the representation learned by the encoder from each EEG token and handle a different number of output tokens, we use an adaptive pooling layer to aggregate information from all EEG representations and add a fully connected layer as a classifier to classify $\mathbf{z}$ into driving intention $\hat{c}$. 

\paragraph{Training MEM with Scheduled Masking Ratio}
The training process commences with a scheduled masking strategy that systematically increases the masking ratio of EEG tokens over time. This incremental elevation of the masking ratio serves the purpose of dynamically challenging the model with varying degrees of information loss, thereby presenting a more diverse and progressively demanding training set. In our experimental setup, the MEM is initially trained with a masking ratio of $0.05$ for $200$ epochs. Subsequently, we iteratively raise the masking ratio in increments of $0.1$, reaching up to $0.9$, every $200$ training epoch. 

We simultaneously optimize both the classification and reconstruction objectives for training MEM. The classification objective is a cross-entropy loss function as written in Eq. \ref{eq:loss-cls}.
\begin{equation}
\label{eq:loss-cls}
    \mathcal{L}_{cls} = -\frac{1}{M}\sum^M_{i=1}c_i\log(p(\hat{c}_i|\mathbf{w}_i))
\end{equation}
, where $M$ denotes the number of training samples, $c_i$ and $\hat{c}_i$ denote the grounth-truth and predicted intention label of the $i^{th}$ sample respectively. The MEM employs a mean square error (MSE) reconstructive objective (Eq. \ref{eq:mse}), measuring the difference between the input EEG signal before masking and the reconstructed EEG signal as written below:
\begin{equation}
\label{eq:mse}
    \mathcal{L}_{mse} = -\frac{1}{M}\sum^M_{i=1}(\hat{\mathbf{w}}_i - \mathbf{w}_i)^2
\end{equation}
The composite training loss function is expressed as follows:
\begin{equation}
\label{eq:loss-all}
    \mathcal{L} = \mathcal{L}_{cls} + \alpha \cdot \mathcal{L}_{mse}
\end{equation}
, where represents, and $\alpha$ is the coefficient for the reconstructive term. In our experiments, we set $\alpha = 0.1$. This training strategy enables MEM to jointly optimize both classification and reconstruction objectives while adapting to increasing levels of information masking.

\section{Experiment}
\subsection{Driving dataset}
Our study utilizes the publicly available Sustained Attention Driving (SAD) dataset \cite{cao2019multi} for analyzing brain activities related to driving intentions as well as evaluation of the proposed MEM model. This dataset encompasses EEG recordings from a total of $27$ participants undertaking a 90-minute driving task in a virtual reality (VR) environment.  Participants were instructed to maintain a straight course throughout the study. However, lane-departure events will be randomly initiated, causing the car to drift away from its designated lane. Participants were required to steer the car back to the original lane. The EEG signals were collected at a sampling rate of $f_s = 500 H\!z$ using a wired EEG cap with $30$ EEG channels and $2$ reference channels. In our experiment, a total number of 16 subjects (5, 11, 22, 35, 40, 41, 42, 43, 44, 45, 48, 50, 52, 53, 54, 55) were selected for DIP evaluation due to the relatively consistent brain patterns exhibited by these subjects throughout their experiment.

A trial in this experiment consists of straight proceeding, lane deviation, steering (response), and eventually back to straight proceeding. Deviation onset will be recorded for each drifting event; response onset and response offset will be recorded at the start and the end of the steering respectively. The driving intention was labeled based on the direction of the steering, and the response onset according to the description in Section \ref{Sec:def}. To label the vigilance state of the participant, We follow the settings from the well-established drowsiness detection research \cite{wei2018toward,cui2022compact,cui2022compact,sia2023eeg} and assign each driving epoch into three levels of vigilance states including drowsy, transition, and alert according to the local reaction time (local-RT), which represents the interval between deviation onset and the subject's response onset. An individualized baseline, alert-RT, was established by calculating the $5$th percentile of a subject's local-RTs. A trial was labeled as `alert' if the local-RT was less than $1.5$ times the alert-RT and `drowsy' if it exceeded $2.5$ times the alert-RT. A transition state encompassed trials falling within the two preceding states. This method effectively distinguishes the three vigilance levels of each subject. Finally, we split each vigilance state's data into non-overlapping train, validation, and test ($80\%$, $10\%$, $10\%$). The complete statistics of the dataset are shown in Table \ref{tab:stats}. We use the training set for training our model, and the validation set for model selection. We report DIP performance evaluated on the test set in the experiment sections. 

\begin{table}[]
    \centering
    \caption{Dataset Statistics.\label{tab:stats}}
    \begin{threeparttable}
\begin{tabular}{llll}
\toprule
\begin{tabular}[c]{@{}l@{}}Vigilance\\ Stage\end{tabular} & \begin{tabular}[c]{@{}l@{}}\# Left Turn\\ Intention\end{tabular} & \begin{tabular}[c]{@{}l@{}}\# Right Turn \\ Intention\end{tabular} & \begin{tabular}[c]{@{}l@{}}\# Straight Proceeding\\ Intention\end{tabular} \\ \midrule
Alert                                                     & 5415                                                             & 5879                                                               & 6805                                                            \\
Transition                                                & 2624                                                             & 2559                                                               & 754                                                             \\
Drowsy                                                    & 1723                                                             & 2243                                                               & 913                                                             \\ \midrule
\begin{tabular}[c]{@{}l@{}}Vigilance\\ Stage\end{tabular} & \begin{tabular}[c]{@{}l@{}}\# Training\\ Samples\end{tabular}    & \begin{tabular}[c]{@{}l@{}}\# Validation\\ Samples\end{tabular}    & \begin{tabular}[c]{@{}l@{}}\# Testing\\ Samples\end{tabular}    \\ \midrule
Alert                                                     & 12140                                                            & 1517                                                               & 1524                                                            \\
Transition                                                & 3530                                                             & 440                                                                & 447                                                             \\
Drowsy                                                    & 1738                                                             & 216                                                                & 224                                                             \\ \bottomrule
\end{tabular}    
    \label{tab:statistics}   
    \end{threeparttable}
\end{table}

\subsection{Analysis of Driver's neural activities for driving intentions}
\begin{figure}[!h]
    \centering
    \subfigure[Central-frontal PSD comparison for left, right, and straight proceeding intentions across vigilance states.]{
        \includegraphics[width=\linewidth]{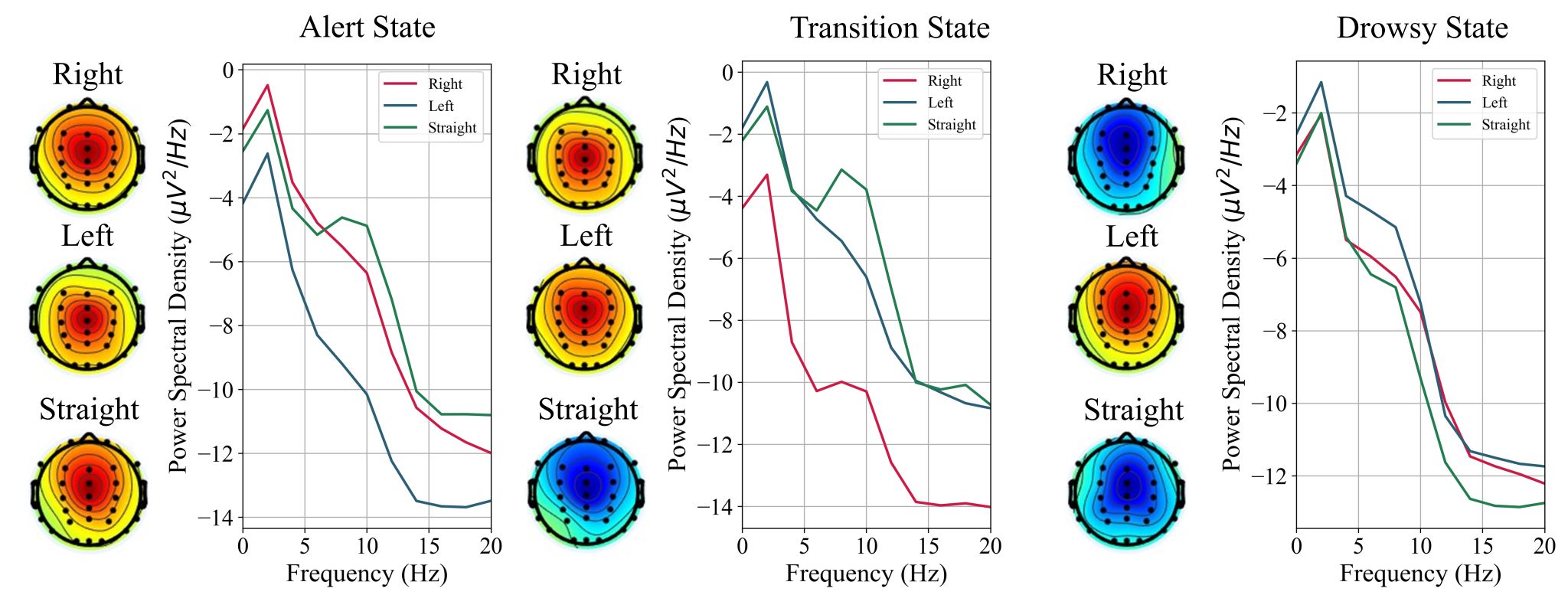}
        \label{fig:ica-frontal}
    }\hspace{0mm}\vspace{0mm}
    \subfigure[Parietal PSD comparison for left, right, and straight proceeding intentions across vigilance states.]{
        \includegraphics[width=\linewidth]{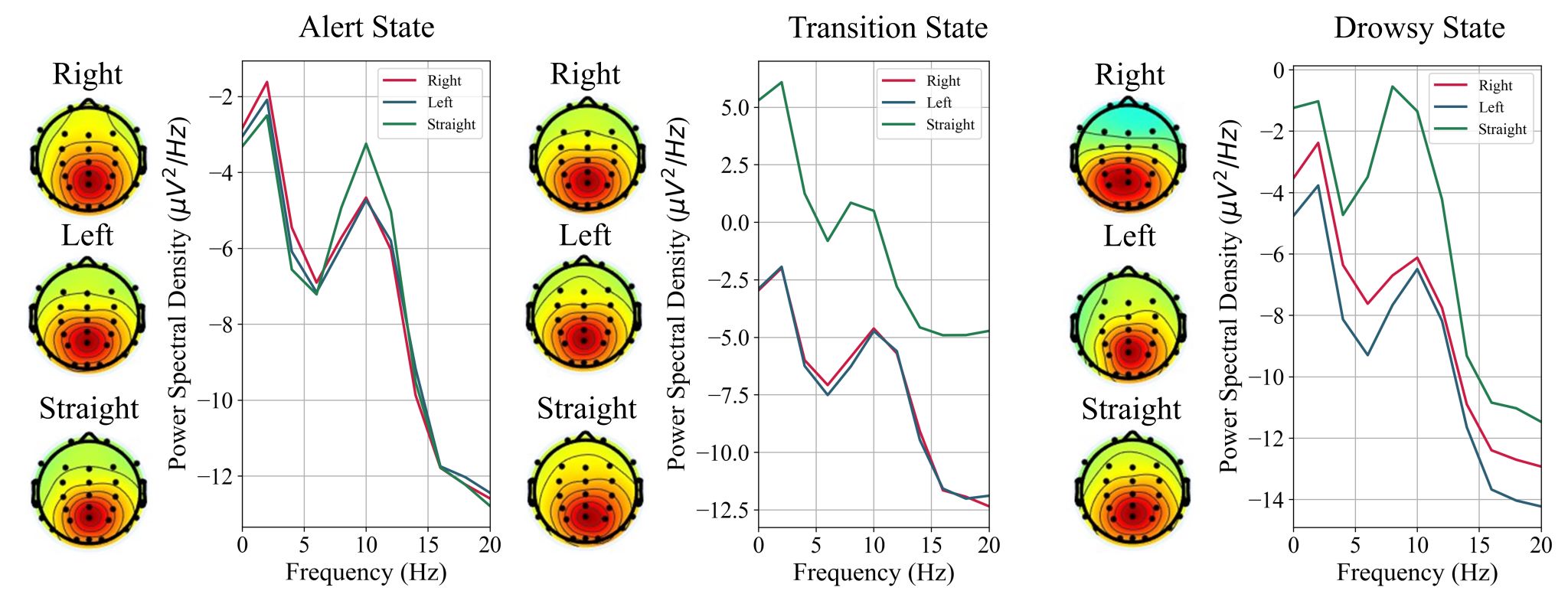}
     \label{fig:ica-parietal}
    }\hspace{0mm}\vspace{0mm}    
    \caption{Independent Components with associated PSD plots from central-frontal and parietal areas.}
    \label{fig:ica-psd}
\end{figure}

To identify highly activated brain sources corresponding to driving intentions, Independent Component Analysis (ICA) is utilized to analyze brain dynamics associated with driving intentions across different vigilance states. We isolate ICs with above $70\%$ probability of representing brain activities for further analysis. Subsequently, we calculate the PSD for each identified brain component to characterize the frequency content of the identified brain sources. Figure \ref{fig:ica-psd} presents PSD plots for driving intentions during different vigilance states, focusing on the central-frontal (Figure \ref{fig:ica-frontal}) and parietal (Figure \ref{fig:ica-parietal}) components for brevity. From a general perspective, ICA decomposition results show that drivers exhibit strong activities originating in the central-frontal and parietal areas, which suggests the involvement of these areas in sensorimotor coordination, cognitive control, and decision-making before steering. This finding is aligned with the neural functionality of the central-frontal and parietal areas \cite{zigmond1999fundamental}. 

For comparison in the central-frontal area (Figure.\ref{fig:ica-frontal}), we could observe that there is a significant spectral difference between left (blue) and right (red) steering intention in the frequency bins from $5$ to $10{H\!z}$ during the three driving intentions. Similarly, a significant difference between straight proceeding intention (green) and turning intention (red and blue) could be observed in the parietal areas (Figure \ref{fig:ica-parietal}). These findings suggest that the driver's driving intention could be discriminated through the brain dynamics. However, the parietal area is frequently less vulnerable to artifacts compared to the frontal area, which enjoys a crucial advantage in the context of driving scenarios where artifacts are prevalent~\cite{webb2021automated}. Furthermore, the parietal region play a more significant role in visuospatial processing, aiding in navigation and responding to changes in the traffic environment. With joint consideration of the neurodynamics and the functionality of the components, we selected $12$ relevant channels as input ($N=12$) to the MEM model, including C3, CZ, C4, CP3, CPZ, CP4, P3, PZ, P4, O1, OZ, O2 covering mainly the parietal area. 

\begin{figure}[!h]
    \centering
    \subfigure[Comparison of Welch's PSD for left, right, and straight proceeding intentions during alert state.]{
        \includegraphics[width=0.8\linewidth]{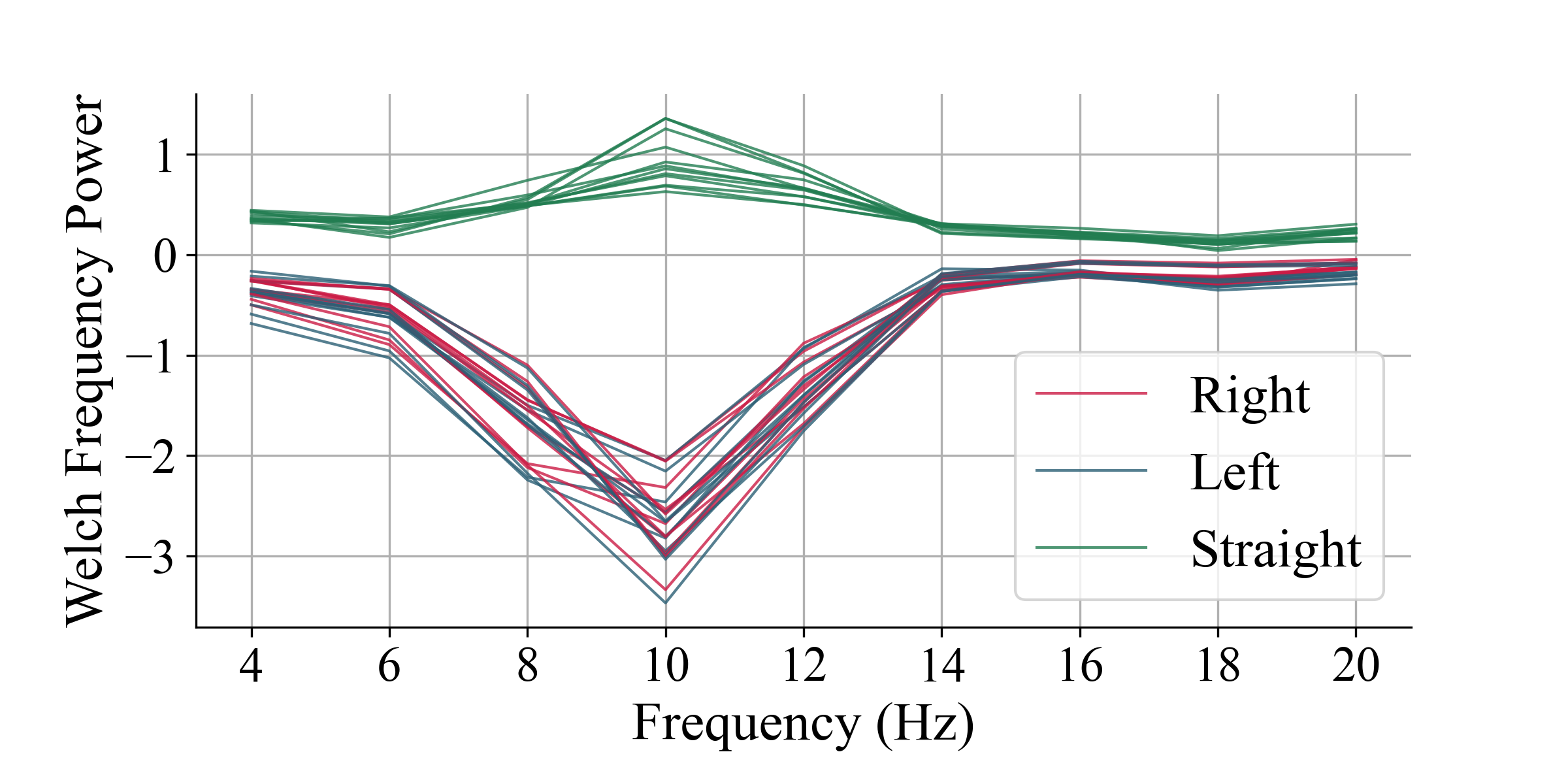}
        \label{fig:welch-alert}
    }\hspace{0mm}\vspace{0mm}
    \subfigure[Comparison of Welch's PSD for left, right, and straight proceeding intentions during transition state.]{
        \includegraphics[width=0.8\linewidth]{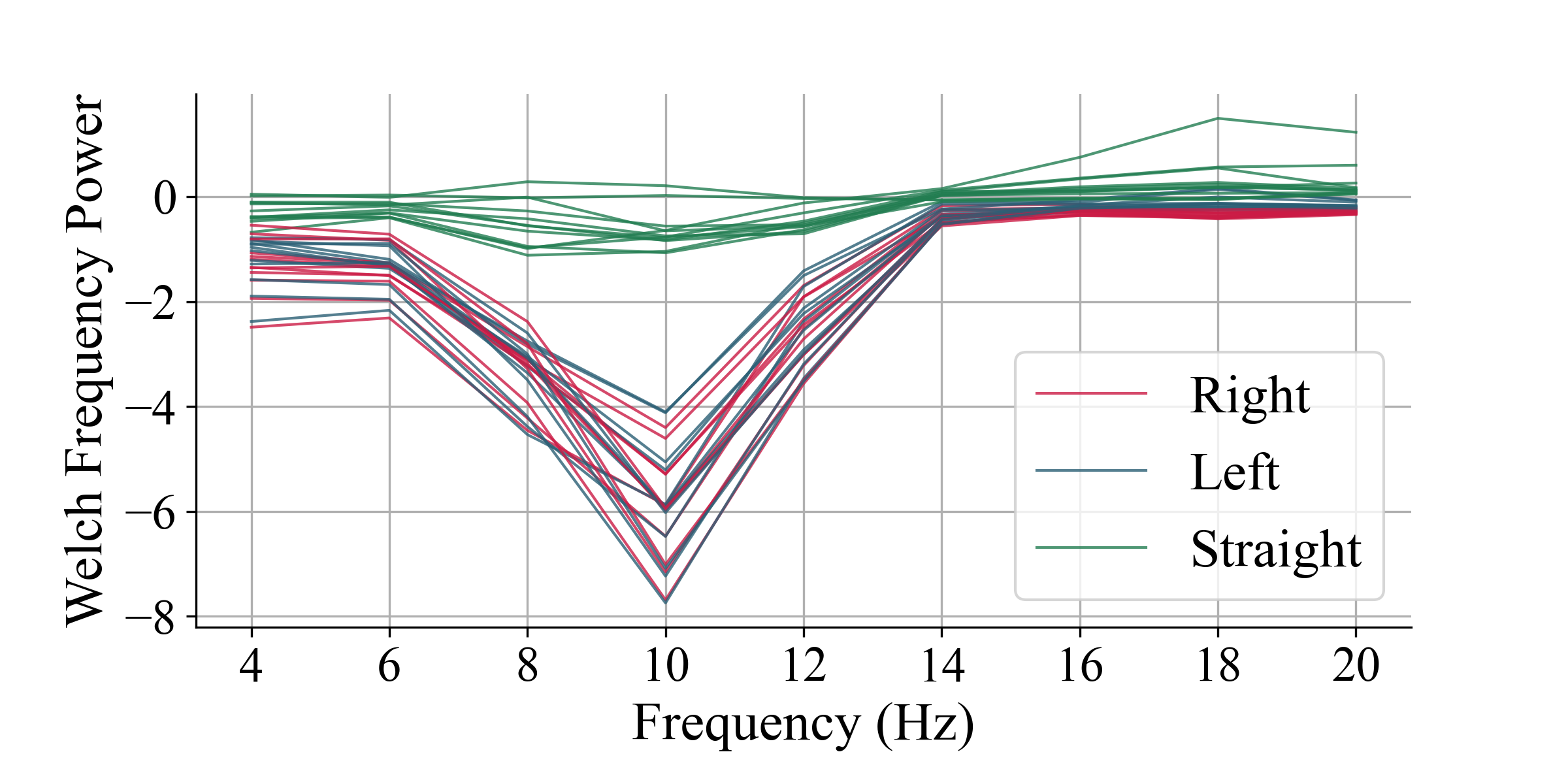}
     \label{fig:welch-transition}
    }\hspace{0mm}\vspace{0mm} 
    \subfigure[Comparison of Welch's PSD for left, right, and straight proceeding intentions during drowsy state.]{
        \includegraphics[width=0.8\linewidth]{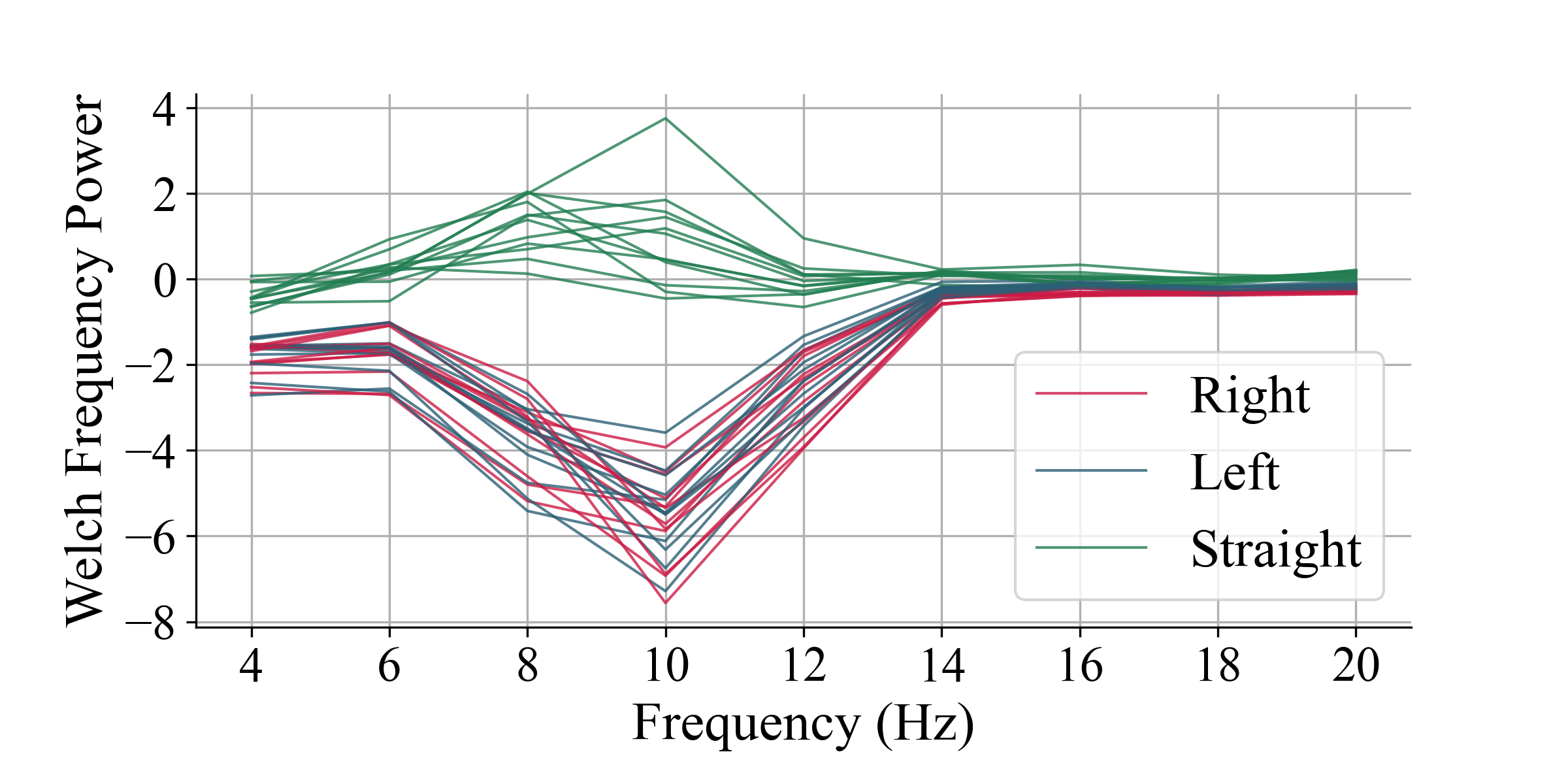}
     \label{fig:welch-drwosy}
    }\hspace{0mm}\vspace{0mm}    
    \caption{Ablation study on multitask learning and effect of our pretrained weights}
    \label{fig:welch}
\end{figure}
While the ICA decomposition reveals clear patterns in EEG signals, its direct application in a real-time decoding system is hindered by computational complexity and limited control over its components. To gain deeper insights into the frequency domain characteristics of brain activity related to driving intentions, we opted for a more lightweight analysis based on the raw EEG waves using Welch's method \cite{welch1967use}. After converting each EEG sample into Welch's frequency domain, we removed reference frequency from the Welch PSD for each subject and aggregated the PSD frequency bands across subjects during left, right, and straight proceeding intentions, we conducted a comprehensive examination. The visual representation of our findings is depicted in Figure \ref{fig:welch}. Notably, across all vigilance states, a substantial difference in the $6$ to $14 H\!z$ frequency bin is observed. This discrepancy is most pronounced in the drowsy state, while it becomes less significant during the transition state. From a neuroscience perspective, this frequency range is commonly associated with the alpha band, known for its involvement in visual processing, attention, and motor planning \cite{zheng2015investigating}. We consider the cause for this discrepancy to be the heightened demands on visual and spatial processing contribute to distinct neural oscillatory patterns within the alpha band especially turning. The deviation from driving straight introduces additional cognitive and motor planning components, further influencing the observed differences in EEG signals. 
Based on the observation of this analysis result and the consideration for increasing adaptability, we select the frequency component from $3$ to $20 H\!z$ as input to the MEM model for learning effective EEG representations. 

\subsection{Evaluation on Driving Intention Prediction}
We first implemented a few baseline methods trained on the SAD dataset's training split. Then we evaluated the baseline methods with our proposed mask modeling framework in Section \ref{sec:method} on various settings.

\paragraph{Baselines}
\begin{itemize}{}{}
\item{\textbf{EEGNet} \quad  A CNN-based baseline with three convolutional blocks, including the temporal convolutional block, the depthwise convolutional, and the separable convolution block \cite{lawhern2018eegnet}. Additionally, it uses a multi-layer perception (MLP) network for classification. The temporal convolutional block employs a series of one-dimensional convolutional layers for capturing both short-term and long-term patterns in the EEG data while the depthwise convolution applies a separate convolutional operation to each input channel, capturing spatial information within individual channels.} To cope with the input shape of our frequency domain EEG data $\mathbf{w}$, we set the kernel size of the temporal and separable convolutional block to $5$. The number of filters is set to $16$ and the depth multiplier is set to $2$. 
\item{\textbf{ViT} \quad A Transformer-based baseline comprises a stack of Transformer blocks \cite{vaswani2017attention}, each consisting of self-attention mechanisms and feedforward neural networks. We maintain the Vit structure to be similar to the proposed MEM encoder using $2$ transformer blocks with $4$ attention head and the hidden size of $512$}
\end{itemize}

\paragraph{Evaluation mectrics}
In our experiment, the performance of the proposed MEM method and the baseline methods are comprehensively evaluated using both the micro accuracy and the marco metrics including precision, recall, and F1-score to provide more balanced measurements of the model's performance across all classes. 

\begin{table*}[!ht]
\centering\setlength\tabcolsep{2.6pt}
\caption{Evaluation result on driving intention prediction\label{tab:main}}
\begin{threeparttable}
\begin{tabular}{c|l|c|cccc|c|cccc|c|cccc}
\toprule
\begin{tabular}[c]{@{}c@{}}Train \\ State\end{tabular}               & Model      & \multicolumn{1}{c|}{\begin{tabular}[c]{@{}c@{}}Test\\ State\end{tabular}} & Accuracy       & Precision      & Recall         & F1             & \begin{tabular}[c]{@{}c@{}}Test\\ State\end{tabular} & Accuracy       & Precision      & Recall         & F1             & \begin{tabular}[c]{@{}c@{}}Test\\ State\end{tabular} & Accuracy       & Precision      & Recall         & F1             \\ \midrule
\multirow{4}{*}{AS}                                                     & EEGNet     & \multirow{4}{*}{AS}                                                      & 45.93          & 42.88          & 41.65          & 34.95          & \multirow{4}{*}{DS}                                  & 46.43          & 38.52          & 38.85          & 38.52          & \multirow{4}{*}{TS}                                  & 43.40          & 42.99          & 47.64          & 37.04          \\
                                                                        & ViT        &                                                                          & 46.06          & 44.10          & 44.32          & 44.07          &                                                      & 47.77          & 26.22          & 42.59          & 32.04          &                                                      & 40.72          & 38.22          & 40.39          & 36.82          \\
                                                                        & MEM$_{w\!/\!o\,{rec}}$&                                                                          & 44.82          & 34.93          & 41.89          & 34.55          &                                                      & 41.96          & 38.40          & 45.67          & 37.84          &                                                      & 42.73          & 40.01          & 41.65          & 39.81          \\
                                                                        & MEM  &                                                                          & 50.85          & 47.84          & 48.05          & 47.12          &                                                      & 48.21          & 26.12          & 42.33          & 31.96          &                                                      & 40.72          & 40.10          & 43.26          & 39.73          \\ \midrule
\multirow{4}{*}{\begin{tabular}[c]{@{}c@{}}AS\\ +DS\\ +TS\end{tabular}} & EEGNet     & \multirow{4}{*}{AS}                                                      & 45.87          & 44.19          & 44.32          & 39.95          & \multirow{4}{*}{DS}                                  & 50.00          & 40.59          & 44.26          & 37.01          & \multirow{4}{*}{TS}                                  & 46.31          & 45.72          & 46.89          & 39.12          \\
                                                                        & ViT        &                                                                          & 59.97          & 59.32          & 59.34          & 59.26          &                                                      & 77.23          & 72.05          & 76.81          & 73.05          &                                                      & 62.19          & 60.30          & 61.44          & 60.00          \\
                                                                        & MEM$_{w\!/\!o\,{rec}}$&                                                                          & 62.14          & 61.99          & 62.04          & 61.83          &                                                      & 78.57          & 73.23          & 76.49          & 74.62          &                                                      & 62.73          & 59.59          & 60.79          & 60.08          \\
                                                                        & MEM  &                                                                          & \textbf{74.48} & \textbf{75.34} & \textbf{73.29} & \textbf{73.78} &                                                      & \textbf{85.19} & \textbf{83.92} & \textbf{86.67} & \textbf{84.77} &                                                      & \textbf{72.04} & \textbf{69.33} & \textbf{73.78} & \textbf{70.64} \\ \bottomrule

\end{tabular}
\begin{tablenotes}
\item[1] AS: Alert State, TS: Transition State, DS: Drowsy State.
\item[2] MEM$_{w\!/\!o\,{rec}}$ denotes MEM without the reconstruction term. 
\end{tablenotes}    
\end{threeparttable}
\end{table*}

\begin{figure}[h!]
    \centering
    \subfigure[Alert State]{
        \includegraphics[width=0.27\linewidth]{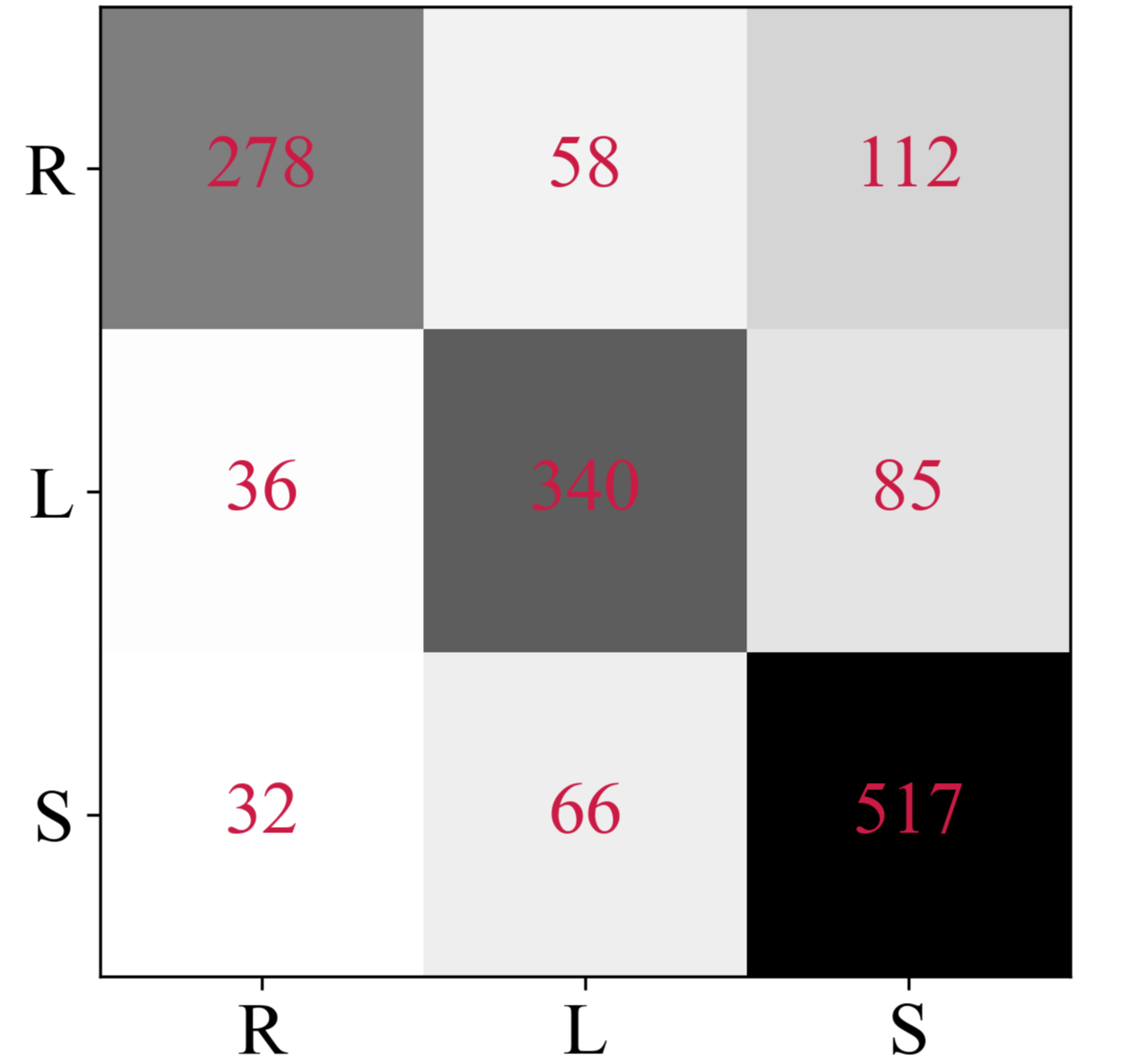}
        \label{fig:cm-alert}
    }\hspace{0mm}\vspace{0mm}
    \subfigure[Transition State]{
        \includegraphics[width=0.269\linewidth]{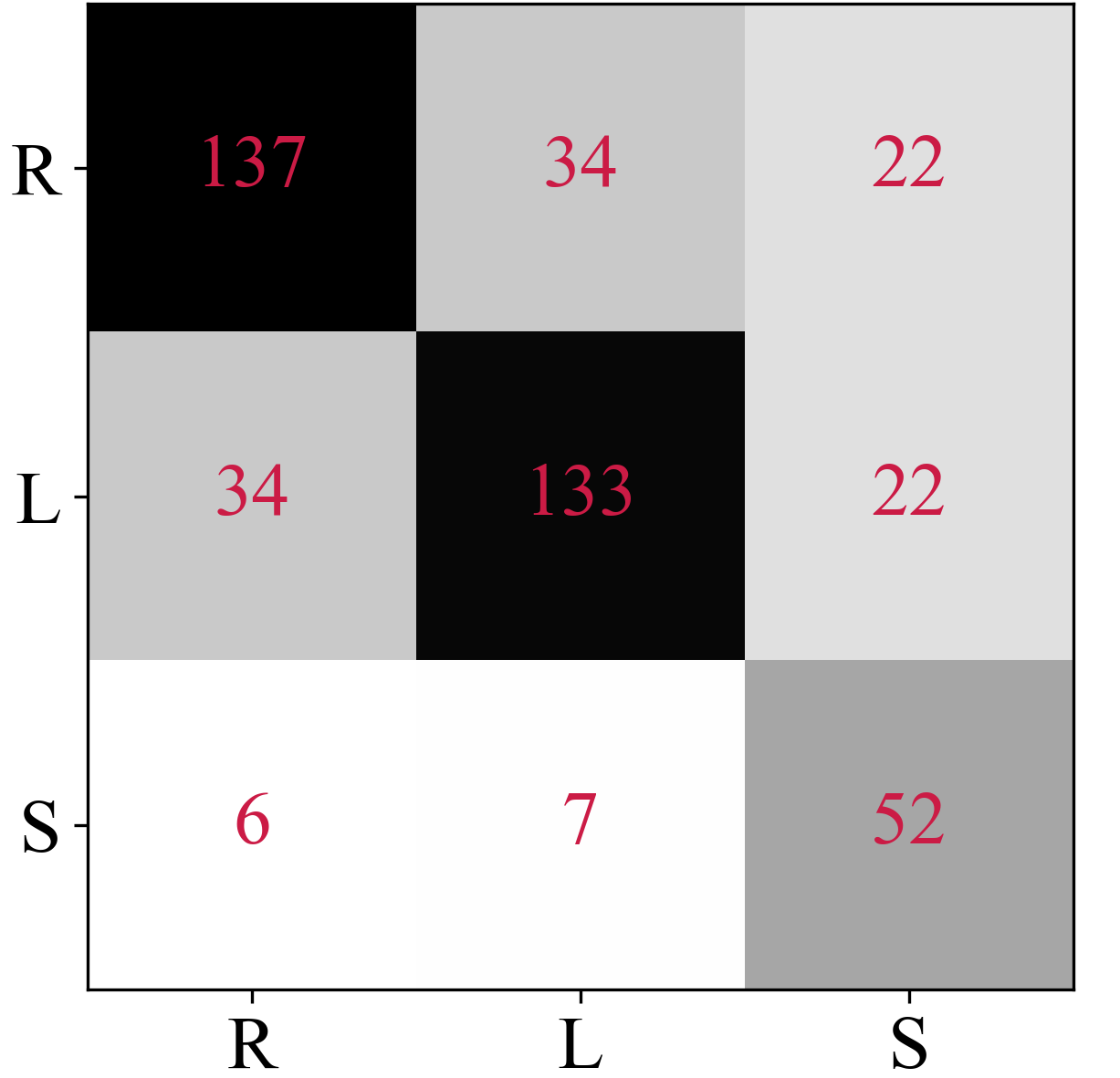}
     \label{fig:confusion-transition}
    }\hspace{0mm}\vspace{0mm} 
    \subfigure[Drowsy State]{
        \includegraphics[width=0.27\linewidth]{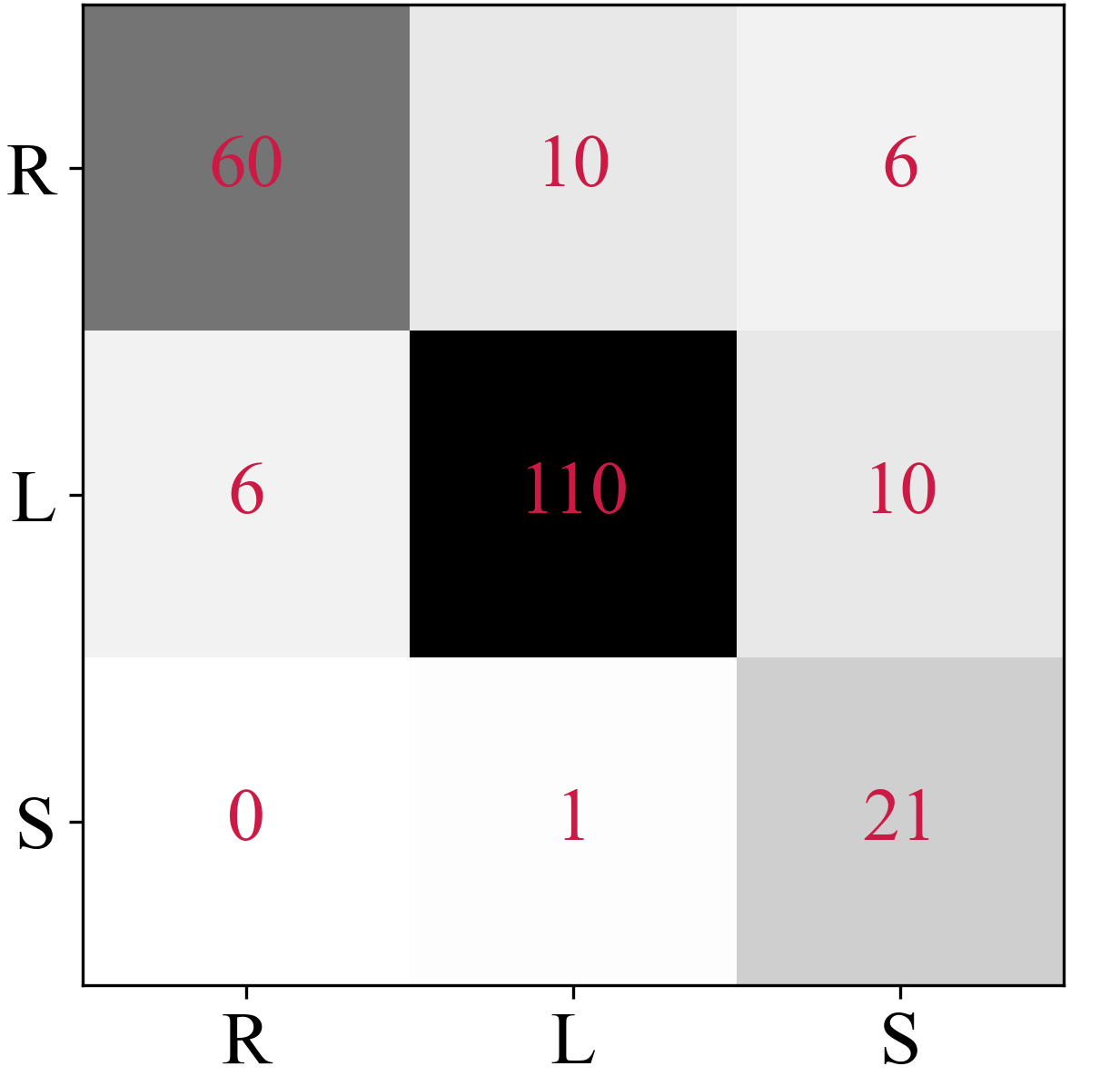}
     \label{fig:confusion-drwosy}
    }\hspace{0mm}\vspace{0mm}    
    \caption{Confusion matrices of MEM model on different vigilance states. R: right turning intention, L: left turning intention, S: straight proceeding intention.}
    \label{fig:confusion-matrix}
\end{figure}

\paragraph{Main Results on DIP}
We first conduct experiments using only the alert state data and then we evaluate the alert (AS), transition (TS), and drowsy (DS) states in a within-subject setting. This is because alert state data is easiest to collect in real-life scenarios. Then we expand the training dataset by adding the other vigilance states (AS+TS+DS), with a hypothesis that models should be able to benefit from the expansion of the training data and the increase of data diversity. Results in Table \ref{tab:main} show the performance of baseline models and our model on predicting driving intentions. Compared to training solely on the alert state driving data, the proposed model is able to benefit from the increase of training samples and vigilance states diversity, exemplified by the significant increase of testing accuracy for the prediction in the transition state (TS) from $38.48\%$ to $72.04\%$. On the other hand, we find that our model's prediction performance in a drowsy state (Accuracy: $85.19\%$) surpasses the performance in an alert state (Accuracy: $74.48\%$) by over $10\%$. We consider the major reason to be due to the additional cognitive load and motor planning in the preparation of a turning maneuver when the driver is forced to respond to deviation events from a drowsy mental state. Thus eliciting stronger and more identical neural activities whereas during an alert state, the brain only exhibits a baseline level of activity necessary for maintaining attention and responsiveness to the change of environment. Figure \ref{fig:confusion-matrix} shows the confusion matrix of the frequency mask model in different vigilance states. For the transition and drowsy vigilance state, the proposed model performs very well in the separation of the straight proceeding state as compared to the right or left turning intention with only a handful of straight samples classified as turning intention. This is especially important for a safety system since undesired turning on the highway could be dangerous. On the other hand, during an alert state, our model excels in classifying between left and right turning intentions, which demonstrates the superior representation learning capacity to exploit spatial and temporal dependencies. 

When compared with other models, our model performs similarly to the EEGNet model when only alert state data is used during training. However, given more data (AS+DS+TS), our model significantly outperforms the EEGNet by a large margin. We mainly contribute to the improvement of the transformer architecture of the proposed method where the transformer building block could be a better memory sponge that could benefit from more diverse training data and could effectively learn to exploit the spatial dependencies of the frequency domain EEG data while the EEGNet is more suitable for learning temporal and spatial filters from raw EEG waves and could not perform as well for the frequency domain signal.   

Compared with transformer-based models (ViT), we observed that the introduction of masking for the input EEG tokens, e.g., MEM and MEM w/o reconstruction, has a positive impact on the prediction performance compared to ViT which is trained on complete EEG information. This is because the masking mechanism introduces additional variations to the input and creates more diverse training conditions for the model. This could help to prevent overfitting to a certain extent, contributing to the improvement of performance in our experiment. On the other hand, thanks to the reconstructive objective during the MEM training, our model is able to learn EEG representations that contain comprehensive information of the original input EEG signal, therefore it has a better understanding of the EEG signal which explains why it achieves the best performance. 

\subsection{Evaluation results on different masking strategies}
\begin{table}[]
\centering
\setlength\tabcolsep{4.5pt} 
\begin{threeparttable}
\caption{Comparison on Different Masking Strategy\label{tab:mask-strategy}}
\begin{tabular}{c|c|c|cccc}
\toprule
\begin{tabular}[c]{@{}c@{}}Training \\ State\end{tabular}               & \begin{tabular}[c]{@{}c@{}}Test\\ State\end{tabular} & \begin{tabular}[c]{@{}c@{}}Mask\\ Strategy\end{tabular} & Accuracy & Precision & Recall & F1    \\ \midrule
\multirow{6}{*}{AS}                                                     & AS                                                   & Channel                                                 & 43.83    & 43.13     & 43.41  & 42.96 \\
                                                                        & TS                                                   & Channel                                                & 45.86    & 44.79     & 44.52  & 44.58 \\
                                                                        & DS                                                   & Channel                                                 & 50.89    & 46.95     & 54.66  & 47.74 \\
                                                                        & AS                                                   & Frequency                                               & 50.85    & 47.84     & 48.05  & 47.12 \\
                                                                        & TS                                                   & Frequency                                               & 38.48    & 37.64     & 41.55  & 38.03 \\
                                                                        & DS                                                   & Frequency                                               & 48.21    & 26.12     & 42.33  & 31.96 \\ \midrule
\multirow{6}{*}{\begin{tabular}[c]{@{}c@{}}AS\\ +DS\\ +TS\end{tabular}} & AS                                                   & Channel                                                 & 65.81    & 65.66     & 65.56  & 65.40 \\
                                                                        & TS                                                   & Channel                                                 & 63.53    & 62.30     & 64.51  & 62.20 \\
                                                                        & DS                                                   & Channel                                                 & 82.14    & 78.54     & 82.53  & 80.12 \\
                                                                        & AS                                                   & Frequency                                               & 74.48    & 75.34     & 73.29  & 73.78 \\
                                                                        & TS                                                   & Frequency                                               & 72.04    & 69.33     & 73.78  & 70.64 \\
                                                                        & DS                                                   & Frequency                                               &85.19    & 83.92     & 86.67  & 84.77 \\ \bottomrule
\end{tabular}
\begin{tablenotes}
\item[1] AS: Alert State, TS: Transition State, DS: Drowsy State.
\end{tablenotes}   
\end{threeparttable}
\end{table}
Table \ref{tab:mask-strategy} illustrates the comparison between two proposed mask modeling strategies for mask EEG modeling. From the results, we could observe that when only trained on alert state data, the MEM model with frequency masking strategy performs better on the seen vigilance states during training (awake) than unseen vigilance states (drowsy and transition), this is exemplified by the observation that the frequency masking model performs better on the awake vigilance state (accuracy on  AS: $50.85\%$) while underperformed by the channel masking model on all other vigilance states. 

However, when presented with a more diverse vigilance condition during the training phase, the frequency masking strategy results in more comprehension and discrimination of driving intentions across all vigilance states when compared to the channel masking strategy. On the other hand, although the channel masking strategy did not outperform the frequency masking strategy in terms of evaluation metrics, it holds enormous practical value as it grants greater flexibility to the working channel number and is robust against broken or missing channels. We consider that it remains a trade-off between better flexibility in real-life applications and the pursuit of better performance. 
\begin{figure}
    \centering
    \includegraphics[width=1.0\columnwidth]{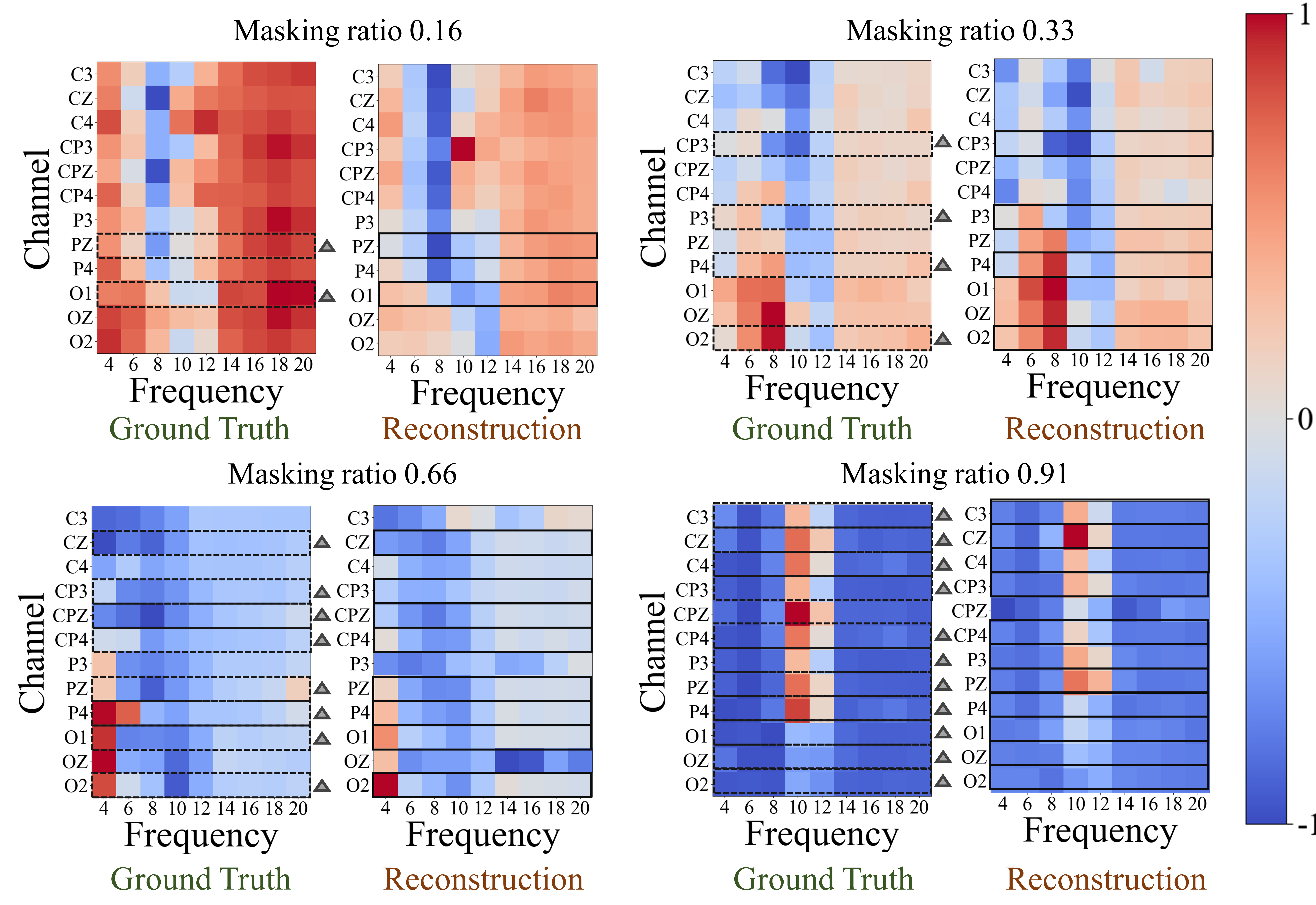}
    \caption{Example reconstruction result on unseen test EEG samples under different masking ratios, best view in color. For each pair, we show the ground-truth frequency spectrogram on the left and the reconstructed spectrogram on the right. The rows with a triangle sign are the masked channels.}
    \label{fig:reconstruct}
\end{figure}
We visualize the reconstructed channels from the channel masking MEM model in different masking ratios to evaluate the model's understanding of the EEG signal in Figure \ref{fig:reconstruct}. As observed from the reconstruction of the unseen EEG signal, the channel MAE could provide a close reconstruction of the masked input channel even if the mask ratio is as high as $0.9$ where $11$ out of $12$ input channels are missing. This indicates the model learns to understand the spatial dependencies of the input EEG channels.

\subsection{Ablation Study}
\begin{figure}
    \centering
    \includegraphics[width=1.0\columnwidth]{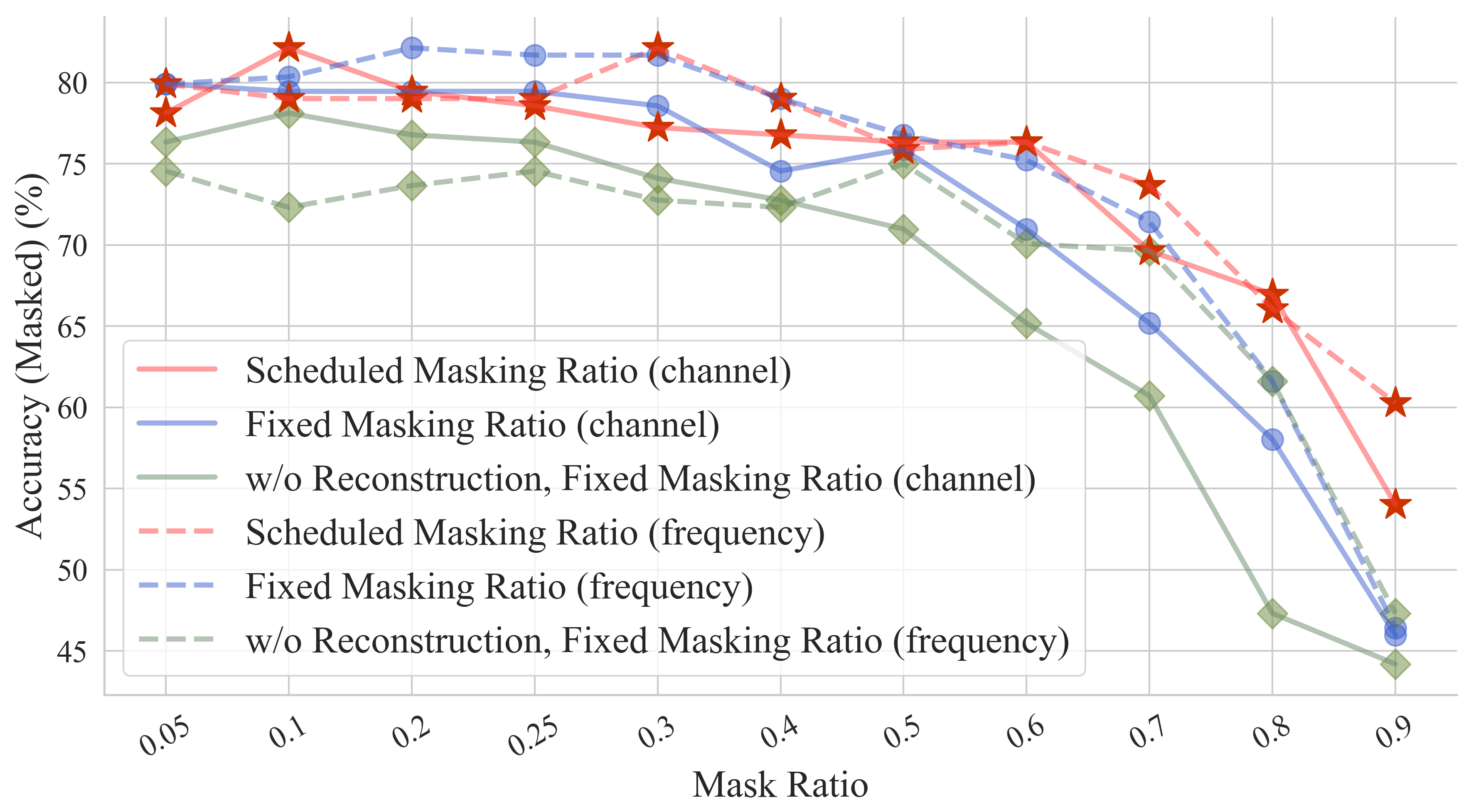}
    \caption{Comparison of the MEM's performance under various masking ratios and training methods.}
    \label{fig:ablation-mask-ratio}
\end{figure}

In this section, we conduct an ablation study on the proposed MEM method, specifically evaluating the impact of the scheduled masking strategy as well as the impact brought by the reconstruction term. We compare three variants of the MEM model including the MEM trained by the scheduled masking ratio (red), MEN trained by a fixed masking ratio (blue), and MEM without reconstructive objective and with a fixed masking ratio (green). We evaluate these models' performance on unseen EEG samples when masks are applied to the input to investigate their resilience against loss of information. For MEM trained by scheduled masking ratio, we report the accuracy achieved in each ratio level during its training process. For other variants, since the mask ratio is fixed for all training epochs, we train a new model for each setting and report the accuracy obtained on the fixed masking ratio. All models are trained using all vigilance states and are evaluated on the drowsy state. The results are visualized in Figure \ref{fig:ablation-mask-ratio}. The comparison highlights the effectiveness of incorporating a reconstructive objective during training (red and blue) compared to the absence of such training (green), leading to improved model understanding of EEG signals and higher prediction accuracy across various masking conditions. The benefits of the reconstructive term become more pronounced, especially when a greater proportion of channels are missing (masking ratio $< 0.5$), underscoring the positive impact of the masked modeling strategy on the DIP task. Furthermore, employing a scheduled masking ratio proves advantageous compared to using a fixed ratio, enhancing the MEM model's adaptability in scenarios with heavy channel disconnection or noise in certain frequency bands. Notably, training the MEM model with a scheduled masking ratio enables it to maintain over $65\%$ prediction accuracy even when $80\%$ of EEG channels are missing or disconnected during driving. Additionally, when information loss is under $60\%$, our model could effectively maintain the prediction accuracy of over $75\%$. This underscores the robustness and adaptability of the MEM model in addressing challenges associated with channel disruptions in real-world driving scenarios.

\section{Conclusion}
This paper presents a pioneering advancement in BCI technology tailored for assisted driving scenarios, with a primary focus on decoding a driver's intentions from EEG signals through the proposed DIP task. The envisioned integration of our task into existing automatic driving systems holds the potential to enable automatic driving systems to align their actions seamlessly with the human's goals or intentions. Our work unveils activation patterns within central-frontal and parietal components, as decomposed by ICA, offering valuable insights into neural dynamics during the intention stage of driving actions. To effectively predict the driver's intention, this paper proposed the MEM framework for learning EEG representations via a masked reconstruction approach. Extensive experiments were conducted on a publicly available dataset, results show that our MEM method achieved $85.19\%$ accuracy on drowsy states and maintained robust performance even with missing or corrupted channels. Emphasizing MEM's adaptability to artifacts and potential applications in real-world scenarios. However, there is room for performance improvement compared to the state-of-the-art speech recognition systems. In the future, we will extend our focus to real-time and continuous decoding of driver intentions without necessitating the pre-segmentation of action intention EEG signals, further enhancing human-machine interaction for safe driving.

\bibliographystyle{ieeetr}
\bibliography{ijcnn}

\end{document}